\documentclass{article}
\usepackage{geometry}
\usepackage[utf8]{inputenc}
\usepackage[T1]{fontenc}
\usepackage{url}

\usepackage{graphicx} 
\usepackage{amsmath}
\usepackage{xcolor}

\urldef{\myurl}\url{https://github.com/AloeUoB/tinyML_indoor_localisation}
\urldef{\byrneurl}\url{https://doi.org/10.6084/m9.figshare.6051794.v1}
\urldef{\ujiurl}\url{https://archive.ics.uci.edu/dataset/310/ujiindoorloc} 

\title{Optimising TinyML with Quantization and Distillation of Transformer and Mamba Models for Indoor Localisation on Edge Devices}

\author{Thanaphon Suwannaphong, Ryan McConville and Ian Craddock \\ \\
 School of Engineering Mathematics and Technology\\
 University of Bristol, UK}
 
\author{
    Thanaphon Suwannaphong\textsuperscript{1,*}, Ferdian Jovan\textsuperscript{2}, Ian Craddock\textsuperscript{1}, Ryan McConville\textsuperscript{1} \\
    \\
    \textsuperscript{1}School of Engineering Mathematics and Technology, University of Bristol, UK \\
    \textsuperscript{2}School of Natural and Computing Sciences, University of Aberdeen, UK \\
}

\date{}

\begin{document}
\maketitle

\begin{abstract}This paper proposes small and efficient machine learning models (TinyML) for resource-constrained edge devices, specifically for on-device indoor localisation. Typical approaches for indoor localisation rely on centralised remote processing of data transmitted from lower powered devices such as wearables. However, there are several benefits for moving this to the edge device itself, including increased battery life, enhanced privacy, reduced latency and lowered operational costs, all of which are key for common applications such as health monitoring. The work focuses on model compression techniques, including quantization and knowledge distillation, to significantly reduce the model size while maintaining high predictive performance.  We base our work on a large state-of-the-art transformer-based model and seek to deploy it within low-power MCUs. We also propose a state-space-based architecture using Mamba as a more compact alternative to the transformer. Our results show that the quantized transformer model performs well within a 64 KB RAM constraint, achieving an effective balance between model size and localisation precision. Additionally, the compact Mamba model has strong performance under even tighter constraints, such as a 32 KB of RAM, without the need for model compression, making it a viable option for more resource-limited environments. We demonstrate that, through our framework, it is feasible to deploy advanced indoor localisation models onto low-power MCUs with restricted memory limitations. The application of these TinyML models in healthcare has the potential to revolutionize patient monitoring by providing accurate, real-time location data while minimizing power consumption, increasing data privacy, improving latency and reducing infrastructure costs.
\end{abstract}

\section*{Keywords}
    TinyML, IoT, Indoor Localisation
    

\section*{Introduction}
Typically, accurate indoor localisation systems rely on a large machine learning models deployed with a centralized remote server using data collected from low-powered battery-based edge devices such as wearables. This requires raw data to be communicated off device leading to increased latency, potential privacy concerns, and higher operational costs for the whole system. Advancements in tiny machine learning (TinyML) have provided a means for implementing advanced machine learning models directly on resource-constrained edge devices. This has potential in various applications including healthcare by enabling more efficient, lower-cost and secure solutions for continuous patient tracking, personalized care and health monitoring.

As the healthcare industry increasingly relies on digital data to optimize patient outcomes, the ability to accurately track and monitor individuals within indoor environments becomes vital. Precise indoor localisation enhances patient monitoring, improves safety, and provides better healthcare outcomes by ensuring timely and accurate data is available to inform medical decisions and interventions \cite{bhamare2024tinyml}. By accurately determining the location of patients, healthcare providers can monitor the elderly or those with cognitive impairments \cite{garcia2023activity} to prevent wandering and ensure timely assistance in emergencies \cite{garcia2022survey}. This technology also facilitates efficient asset tracking \cite{avellaneda2023tinyml}, optimizing the use of medical equipment and reducing operational inefficiencies. Moreover, indoor localisation supports personalized healthcare \cite{poyiadzi2020detecting} and after treatment monitoring \cite{McConville2021} by collecting detailed movement patterns that can inform tailored interventions and treatments. Implementing such systems on resource-constrained devices through TinyML models ensures that these benefits can be realised with minimal infrastructure costs and maximal privacy protection, as data processing occurs locally on the device. 

On-device indoor localisation offers several benefits, including enhanced privacy and security, as sensitive location data is processed locally rather than transmitted over networks. This approach reduces latency and improves real-time tracking accuracy, which is crucial for applications like patient monitoring in healthcare. Additionally, it decreases reliance on external infrastructure, lowering costs and increasing system robustness and reliability in diverse environments. On-device processing also enables smart battery management, reducing power consumption by putting the device into sleep mode during periods of inactivity, such as when users are stationary indoors, asleep in their bedrooms, or outside the monitoring area. This smart system extends the operational longevity of monitoring devices per charge cycle.  Thus, on-edge indoor localisation  is a valuable technology supporting the seamless integration of digital health solutions into everyday patient care routines.

Developing efficient models for indoor localisation on resource-constrained devices presents significant challenges. Low-power micro-controller units (MCUs) have limited memory, processing power, and battery life, which restrict the size and complexity of deployable models. TinyML, the field focused on deploying machine learning models on resource-constrained devices \cite{kulkarni2024tinyml}, is crucial for overcoming these limitations. TinyML enables the execution of machine learning algorithms on small, low-power devices, making it possible to implement advanced functionalities like indoor localisation while adhering to stringent resource constraints typical of wearable devices and IoT sensors.

Model compression techniques offer one approach for developing TinyML models, as they help reduce the size and computational requirements of machine learning models without significantly compromising performance. This paper focuses on two specific compression techniques: quantization and knowledge distillation. Quantization reduces the precision of the model's weights and activations, effectively decreasing the model size and improving inference speed \cite{gholami2022survey}.  Knowledge distillation involves transfering knowledge from a large, complex model (teacher) to a smaller, simpler model (student), achieving comparable accuracy with fewer resources \cite{hinton2015distilling}. These techniques are proven effective in various fields and hold significant potential for enhancing on-device indoor localisation.

In proposing the appropriate MCU from available market options, our focus has been on minimizing energy consumption to extend monitoring periods, while ensuring sufficient memory capacity to load and execute the indoor localisation model entirely within the MCU’s RAM. Table \ref{tab:TargerMCU_list} shows that devices with 64KB RAM or less tend to have particularly low active-mode current, making them more suitable, such as the ADuCM3029, RL78/G13, CC2650 and STM32L4. Therefore, this study aims to develop a tiny indoor localisation model with size less than 64 KB.

\begin{table}
\footnotesize
    \centering
    \caption{Examples of low-power MCUs suitable for wearable devices.}
    \label{tab:TargerMCU_list}   \begin{tabular}{|c|c|c|c|} \hline 
         MCU names&  RAM (KB)&  Flash (KB)& Active-mode current\\ \hline 
 \multicolumn{4}{|c|}{2 KB RAM or less}\\\hline 
         MSP430FR5969&  2&  64& 1.6mA
\\ \hline 
         ATmega328P&  2&  32& 1.5mA
\\ \hline 
 \multicolumn{4}{|c|}{2 to 64 KB RAM}\\ \hline 
         CC2650&  20&  128& 2.9mA
\\ \hline 
 RL78/G13& 32& 512&2.11mA\\ \hline 
         ADuCM3029&  64&  128& 0.96mA
\\ \hline 
         STM32L4&  64&  256& 4.0mA
\\ \hline 
 \multicolumn{4}{|c|}{More than 128 KB RAM}\\ \hline 
         MAX32630&  512&  2048& 10.1mA
\\ \hline 
         EFM32&  128&  1024& 10.5mA
\\ \hline
    \end{tabular}
    
\end{table}

The primary objectives of this paper are to develop a highly efficient, small-scale indoor localisation model that fits within the memory constraints of low-power MCUs and to provide a thorough evaluation of quantization and knowledge distillation as model compression techniques to achieve this goal. We aim to evaluate the performance of a state-of-the-art transformer model and a Mamba-based model for indoor localisation, applying the aforementioned compression techniques to optimize their sizes for deployment on resource-constrained devices.

This provides insight into the most effective strategies for achieving efficient on-device localisation within the limitations of low-power MCUs, with the aim of advancing the development of robust and energy-efficient healthcare monitoring systems that seamlessly integrate into users' daily routines. The main contributions of the paper are as follows:
\begin{itemize}
  \item We propose a small and efficient model for indoor localisation suitable for resource-constrained devices by compressing state-of-the-art transformer-based models.
  \item We are the first to propose Mamba based architectures for indoor localisation, demonstrating that even without model compression, it is suitable for devices with less than 32 KB of memory.
  \item We systematically evaluate indoor localisation models for in-home datasets and large, multi-building datasets, all with model sizes under 64 KB, suitable for deployment on low-power MCUs, providing benchmarks to guide model selection based on hardware constraints.
\end{itemize}

\section*{Related Work}
Indoor localisation has seen significant advancements with the development of high-performing models that leverage complex algorithms and extensive computational resources \cite{mendez2024machine}. Models such as those based on deep learning have shown great accuracy and robustness in determining precise indoor locations \cite{jovan2023multimodal}. However, a major limitation of these state-of-the-art models is their size and computational demand, making them unsuitable for deployment on edge devices with limited resources. The requirement for high computational power and large memory footprints restricts their application in real-time, on-device scenarios.

Tiny Machine Learning (TinyML) has emerged as a promising solution to address the challenges of deploying machine learning models on resource-constrained devices \cite{ray2022review}. TinyML techniques aim to reduce the size and computational requirements of models while maintaining their performance. This approach is particularly beneficial for applications like indoor localisation, particularly for healthcare purpose, where real-time processing and low power consumption are critical.

Model compression techniques like quantization, pruning, low-rank factorization, and knowledge distillation are designed to reduce model size, each with its own set of limitations. The summarised  comparison of these model compression techniques is presented in Table \ref{tab:compression_comparison}. Pruning, which removes less significant weights, can create sparse models that are harder to optimize and may lose accuracy due to the removal of crucial connections \cite{zhu2017prune}. Moreover, it often requires specialized hardware to support sparse operations, making it impractical for low-power micro-controllers (MCUs). Low-rank factorization, which approximates weight matrices with lower-rank versions, can struggle to retain the model's expressiveness, particularly in complex tasks, leading to potential performance drops \cite{sainath2013low}. This technique is typically more effective in large models with very high performance, rather than in smaller models like those used in this study. In contrast, quantization and knowledge distillation offer more practical solutions for TinyML applications, as they do not require specialized hardware and are more effective at reducing the size of smaller models while maintaining good performance.

\begin{table}[ht]
\centering
\caption{Comparison of Model Compression Techniques for TinyML Applications}
\label{tab:compression_comparison}
\begin{tabular}{|l|l|l|l|l|}
\hline
\textbf{Characteristic}      & \textbf{Quantization} & \textbf{KD} & \textbf{Pruning} & \textbf{Low-Rank Factor} \\ \hline
\textbf{Memory Reduction}     & High      & High                             & High (Sparsity)  & Moderate                        \\ \hline
\textbf{Hardware Requirements} & None                  & None                             & Require  & None                            \\ \hline
\textbf{Ease of Implementation} & Easy                  & Moderate                         & Moderate         & Complex                         \\ \hline
\textbf{Performance Preservation} & Moderate           & High                             & Moderate         & Low                             \\ \hline
\textbf{Suitability for TinyML}  & High                 & High                             & Low              & Low                             \\ \hline
\end{tabular}
\end{table}

Quantization and knowledge distillation are among the most popular TinyML techniques for model compression \cite{abadade2023comprehensive}. Quantization reduces model size and computational load by decreasing the precision of the numbers representing the model’s parameters, which can instantly reduce the model's size by about half \cite{polino2018model}. Its straightforward application allows for easy implementation, especially when compressing already small and efficient models for deployment on resource-constrained devices. Knowledge distillation, on the other hand, involves transferring knowledge from a larger model (the teacher) to a smaller one (the student) by training the student to mimic the behaviour of the teacher \cite{hinton2015distilling}. Although this process requires additional training, it offers flexibility in the size and architecture of the models, allowing the teacher to be a complex, high-performing model while the student remains a tiny, simple neural network. This flexibility is crucial for developing models that meet the stringent constraints of TinyML applications. Thus, we investigate these two techniques to achieve small and efficient indoor localisation models aimed to operate in low-power MCU devices.

There have been several attempts to apply these techniques to compress indoor localisation models for edge devices while maintaining accuracy. For instance, studies have applied quantization to develop tiny indoor localisation models on MCU devices using various models such as DNN \cite{hayajneh2022channel,kotrotsios2022design, girolami2023tinyml, avellaneda2023tinyml}, LSTM \cite{girolami2023tinyml}, and one-layer MLP \cite{jones2023tiny}. Similarly, some studies have evaluated the potential of knowledge distillation to develop on-device indoor localisation models using simple neural network models such as CNN \cite{mazlan2022fast, mazlan2022teacher}, and DNN \cite{putrada2023knowledge, al2024knowledge}. While these studies successfully developed compact and efficient models for indoor localisation, they primarily used simple neural network models not specifically designed for sequence modeling tasks like RSSI localisation.

The Transformer model \cite{vaswani2017attention}, with its core attention layer, has been highly successful in sequence modeling due to its effectiveness in handling information-dense data within a context window. This capability makes it suitable for modeling complex data. Transformers have been widely adopted in various fields, particularly in natural language processing and sequence data analysis \cite{lin2022survey}. Notably, the transformer-based model, called MDCSA, has achieved state-of-the-art performance in indoor localisation, particularly when tested on complex, real-world datasets \cite{jovan2023multimodal}. MDCSA outperformed other leading models in time-series data tasks, as evidenced in its comparative analysis table. Given their high performance, Transformers have significant potential for developing efficient models for on-device localisation. Therefore, our study uses MDCSA as a strong baseline.

Another recent advancement in machine learning for sequence modeling that potentially benefits the development of TinyML models is Mamba \cite{gu2023mamba}, a novel class of structured state space models (SSMs). Mamba is competitive with the well-known transformer model as it combines the strengths of RNNs and CNNs to efficiently handle sequence modeling with linear scaling in sequence length. Mamba is also lightweight compared to transformers due to its SSM-based structure, which offers potential for on-device applications, particularly on resource-constrained MCU devices. It is a SoTA architecture for time-series tasks, with proven performance in various applications. For instance, Ahamed et al., \cite{ahamed2024timemachine} showcases Mamba’s efficiency, even in resource-constrained settings. To the best of our knowledge, Mamba has not been previously studied within the context of TinyML for indoor localisation. This presents a unique opportunity for our research to explore the potential of Mamba in developing an efficient, compact model for indoor localisation using RSSI data.

In this study, we propose a Mamba-based architecture tailored for indoor localisation and apply model compression techniques, comparing its performance with MDCSA, the state-of-the-art transformer-based indoor localisation model. This work addresses a gap in the TinyML field, where most existing studies focus on simpler neural networks like CNNs and DNNs. By exploring advanced machine learning models such as Mamba and Transformers, we aim to utilise their capabilities to improve the efficiency and accuracy of indoor localisation models for resource-constrained devices.

In summary, our methodology will involve Mamba and Transformer models for on-device indoor localisation using quantization, knowledge distillation, and hybrids, to achieve the best model under limited device constraints.

\section*{Datasets}
We evaluate our approaches on two popular public datasets for indoor localisation. One uses BLE RSSI collected from 4 residential homes \cite{byrne2018residential}. Another dataset, called UJIindoorloc, uses WIFI RSSI collected from 3 university buildings \cite{torres2014ujiindoorloc}. Together they provide coverage over two common types of indoor localisation, each with unique challenges. The summarised details of these datasets are presented in Table \ref{tab: dataset}.

\begin{table}[h!]
\centering
\caption{Summary of Dataset Details}
\label{tab: dataset}
\begin{tabular}{@{}llcccccc@{}}
\hline
   \textbf{Name} & \textbf{Building Type} & \textbf{No. of} & \textbf{No. of} & \textbf{scripted} & \textbf{unscripted} & \textbf{RSSI} \\ 
     &  & \textbf{APs} & \textbf{Classes} & \textbf{Data} & \textbf{Data} & \textbf{Signal} \\ \hline
House A                  & 1-bedroom apartment & 8                  & 4                       & 39 mins                   & 49.6 mins                & BLE                 \\
House B                  & 2-bedroom, 2-floor house & 11                 & 11                      & 117.8 mins                & 47.2 mins                & BLE                 \\
House C                  & 3-bedroom, 2-floor house & 11                 & 9                       & 82 mins                   & 237 mins                 & BLE                 \\
House D                  & 2-bedroom, 2-floor house & 11                 & 10                      & 97 mins                   & 178.4 mins               & BLE                 \\
UJIIndoorLoc             & 3 multiple-floors building & 520                & 59                      & 19,937 samples            & 1,111 samples            & Wi-Fi               \\ \hline
\end{tabular}
\end{table}

\subsection*{In-Home Localisation Dataset}
The in-home localisation dataset is a public dataset collected from four residential houses and uses BLE signals for indoor localisation \cite{byrne2018residential}. This dataset serves the purpose of indoor positioning systems within home environments which is increasingly common for healthcare purposes \cite{farahsari2022survey}.

The dataset includes four houses: House A (one-bedroom apartment), Houses B and D (two-bedroom, two-floor houses), and House C (the largest, three-bedroom, two-floor house). Each area within the houses is labelled at the room level, with some larger rooms classified into multiple classes (e.g., Living Area A and Living Area B). Additional classes include hallways, stairs, and outside gardens. House A, B, C, and D contain 4, 11, 9, and 10 classes, respectively.

The dataset employs wrist-worn accelerometers transmitting BLE signals at 5Hz. Raspberry Pi-based access points (APs) receive these signals and record RSSI values from the person wearing the device. The RSSI values range from -110 to 0, indicating signal strength. Each room is equipped with at least one AP for room-level localisation. Larger rooms, however, may have multiple APs strategically placed to ensure accurate signal reception throughout the space.

Data collection utilized an automated system incorporating binary floor tags placed at one-meter intervals and a chest-strapped camera capturing images as participants moved throughout the houses. This method provided precise location labels.

The dataset is organized into two experiments: \textit{fingerprint}  and \textit{free-living}. In the fingerprint experiments, controlled experiments were conducted where participants followed scripted paths to systematically visit every area within the houses. This method ensured comprehensive coverage for training and validation purposes. Conversely, the Free-living section captured participants' daily activities within the residences, reflecting their natural movement patterns and interactions with the environment. Houses A, B, C, and D have scripted fingerprint measurements of 39, 117.8, 82.0, and 71.0 minutes. The corresponding unscripted living recordings are 49.6, 47.2, 237.0, and 178.4 minutes.

For model training and validation, 75 percent of the fingerprint data is used to train the model, with the remaining 25 percent used for validation. The entire free-living dataset is reserved for testing purposes, providing real-world scenarios and unseen data to evaluate the trained model's performance.

\subsection*{UJIIndoorLoc Dataset}
The UJIIndoorLoc dataset \cite{torres2014ujiindoorloc} was collected for evaluation of indoor positioning systems using WLAN/WiFi fingerprinting in larger spaces. It covers multiple buildings and floors at Jaume I University (UJI), serving as a reliable testbed for developing and validating precise multiple building indoor localisation models. We use this dataset to extend our evaluation beyond in-home localisation, demonstrating the efficacy of our approach across larger and multiple buildings, rather than limited to single home environments.

The UJIIndoorLoc database contains three buildings within the university, each with multiple floors. Two of the buildings have four floors each, while the third building comprises five floors. The dataset includes location information in terms of longitude, latitude, floor number, and building ID, facilitating the prediction of the exact position within the building.

The dataset was collected using 25 different Android devices, which acted as the signal transmitters. The receivers were 520 wireless access points (APs) scattered throughout the buildings. The signal type used in this dataset is WiFi, with the signal frequency ranging from 0.1 Hz to 1 Hz due to varying user activities and device usage. Signal strength is recorded as negative integer values, ranging from -104 dBm (indicating an extremely weak signal) to 0 dBm. A positive value of 100 is used to indicate instances where a wireless access point was not detected.

Data for the UJIIndoorLoc dataset were collected from more than 20 different users using two Android applications: CaptureLoc and ValidationLoc. These applications interfaced with reference map services published on an ArcGIS server, which provided detailed geographic information of the building interiors and training reference points. For the training set, 18 users captured RSSI signals from 924 reference points using the CaptureLoc application, resulting in a dataset with 19,937 samples. The validation set involved 14 users who collected data for approximately 20 minutes in each building, capturing WiFi signals randomly which might come from locations not included in the training set, resulting in 1,111 samples.

For model training and validation, 75 percent of the training set was used to train the model, while the remaining 25 percent was used to validate its performance. The validation set, consisting of 1,111 samples, was used as an independent testing set. This approach ensured that the model was tested on previously unseen data, enhancing the robustness and reliability of the localisation predictions.

\subsection*{Data Pre-Processing}
\subsubsection*{In-Home Dataset}
We handle missing values and prepare the dataset for analysis to ensure the integrity and consistency of the data. Missing data are initially addressed using forward filling with a value of the latest timestamp for 1 second given the limited movement capability within a home. Missing data at this rate is typically due to dropped packets by the hardware. Any remaining gaps were represented by -120, a value beyond the feasible range of RSSI readings, to clearly indicate missing entries (such as when the person was out of the home). We also applied a windowing technique with 4-second windows and a 50 percent overlap to capture temporal patterns. Lastly, min-max normalization was used to scale data values between 0 and 1, ensuring uniform feature scaling and facilitating effective model training for room classification.

\subsubsection*{UJIIndoorLoc Dataset}
This dataset has already managed missing data using a placeholder value of 100 to indicate undetected signals, thus removing the need for further handling of missing entries. Given the dataset's irregular sampling rate, ranging from as low as 0.1 Hz to as high as 1 Hz, we opted to forgo windowing techniques to avoid compromising the accuracy of location predictions through upsampling. Min-max normalization was applied to scale the data within a 0 to 1 range, ensuring consistent feature representation. Additionally, we assigned class labels to each area within the building, categorizing data points by specific locations and floors. As a result, the dataset consist of 59 classes throughout all floors of the three buildings. This is done to prepare the dataset for a classification task, as our objective is to classify the data points into distinct areas rather than predicting continuous values. This facilitates the localisation process and enhances the predictive modelling of spatial data. 

\section*{Methodology}
\subsection*{Model Compression Techniques}
We apply two model compression techniques—quantization and knowledge distillation—to develop a compact and efficient indoor localisation model suitable for low-power MCU devices. Quantization is particularly advantageous due to its straightforward application, allowing for easy implementation without additional steps, especially when compressing already small models. Knowledge distillation, while requiring additional training steps, offers greater flexibility in the size and architecture of the models. This approach enables the use of a complex, high-performing teacher model to train a smaller, simpler student model, which is crucial for meeting the constraints of the target device in TinyML applications.

\subsubsection*{Quantization}
Quantization is a technique used to reduce the size and computational requirements of machine learning models, making them suitable for deployment on resource-constrained devices \cite{gholami2022survey}. This method works by decreasing the precision of the numbers used to represent the model’s parameters, effectively reducing the memory footprint and computational complexity. They are seeing increasing use by those deploying Large Language Models (LLMs) to (still) high powered machines \cite{dettmers2022gpt3}. Nonetheless, we will use it here to apply them to tiny devices. 

In a typical machine learning model, parameters are represented using 32-bit floating-point numbers (FP32). Quantization replaces these high-precision numbers with lower-precision alternatives, such as 8-bit integer (int8) representations. This reduction in bit-width translates directly into smaller model sizes and faster computations, as lower-precision arithmetic operations require fewer computational resources.

To convert from floating-point to integer values, the original floating-point values are multiplied by a scaling factor and then rounded to the nearest whole number, as shown in Equation \ref{eq:quantization}.

\begin{equation} \label{eq:quantization}
 Q = \text{round}\left(\frac{x}{\text{scale}} + \text{zero\_point}\right)
\end{equation}

\begin{equation} \label{eq:scale}
 \text{scale} = \frac{\text{max}\_\text{float} - \text{min}\_\text{float}}{\text{max}\_\text{int} - \text{min}\_\text{int}}
\end{equation}

\begin{equation} \label{eq:zero_point}
 \text{zero\_point} = \text{round}\left(-\frac{\text{min}\_\text{float}}{\text{scale}}\right)
\end{equation}

The $scale$ is a floating-point scaling factor for converting values from higher to lower precision. For example, the usual case is converting FP32 values (scale $-3.4 \times 10^{38}$ to $3.4 \times 10^{38}$) to int8 values (scale $0$ to $255$). The $zero\_point$ is an integer value that maps to the real zero in the quantized value range. It adjusts the range of the quantized values to ensure that zero in the floating-point range is accurately represented in the integer range. Various quantization approaches differ in how they determine these quantization parameters.

We apply post-training quantization to develop the TinyML model. The post-training quantization involves training the model with high precision and then converting the weights and/or activations to lower precision after training. This approach is straightforward and can be applied to pre-trained models without requiring access to the original training data, making it appropriate for those who want to take SoTA models and apply them in resource constrained settings. For our task we investigate two main types of post-training quantization: static and dynamic quantization.

\textbf{Static quantization} quantizes the weights and activations of the model during the model conversion process. This method requires a calibration step, where the model is run with a representative dataset to determine the optimal scaling factors for the weights and activations. We apply the LLM.int8() static quantization method \cite{dettmers2022gpt3} to achieve a TinyML model for indoor localisation. The LLM.int8() quantization method improves the performance of large-scale models by handling outlier features with 16-bit floating point (FP16) precision, while quantizing non-outlier values into 8-bit integers (int8) for memory efficiency. The process involves three main steps: extracting outliers, applying vector-wise quantization to non-outliers, and combining mixed-precision results. This approach maintains efficiency while preserving accuracy for extreme values, specifically targeting the quantization of linear layers in transformers where the weights and activations can be extreme.

In our study, we apply the LLM.int8() static quantization to compress our Transformer- and Mamba-based models for indoor localisation into a TinyML-friendly size. The LLM.int8() method is particularly beneficial for our model because it efficiently handles outlier features, which are common in the complex signal environments typical of indoor localisation tasks. By using FP16 precision for these outliers and int8 for other values, we maintain the model's accuracy while significantly reducing its size. For example, it can reduce the size of a linear layer by up to 75 percent, dependent on the number of outliers. We note that this method only works on linear layers, as it was originally designed for large language models that rely heavily on linear layers, making it suitable to apply to our Transformer and Mamba models which mainly consist of multiple linear layers. This approach is critical for ensuring that our models, which includes linear layers in the attention head of the Transformer and structure state space layer of the Mamba, performs reliably even with the extreme variations in signal strength encountered in diverse indoor settings.

\textbf{Dynamic quantization}, in contrast, only quantizes the model weights ahead of time and leaves activations in their original precision during the calibration process. At runtime, activations are dynamically quantized as they pass through the model. This reduces the size of the model weights and speeds up model execution. We use the PyTorch library for post-training dynamic quantization. This method is ideal for situations where model execution time is dominated by loading weights from memory rather than computing matrix multiplications, which is true for Transformer models with small batch sizes, such as the one used in this study.

Dynamic quantization differs from LLM.int8() as it does not treat outliers separately and is a tensor-wise process rather than vector-wise like LLM.int8(). The tensor-wise quantization scheme applies a single scale and zero-point to the entire weight matrix. Each tensor matrix has distinct quantization parameters, allowing each layer to be quantized independently based on its data distribution. This method is simpler and computationally less intensive compared to vector-wise quantization. It is typically used when uniform quantization across the entire weight matrix is sufficient, thus we compare this approach with the LLM.int8() to investigate the optimal technique to develop TinyML for indoor localisation. This dynamic quantization supports \texttt{nn.Linear}, \texttt{nn.LSTM}, \texttt{nn.RNN}, \texttt{nn.GRU}, and \texttt{nn.Embedding}, making it suitable for our models which mainly consist of linear layers.

\subsubsection*{Knowledge Distillation}
Knowledge distillation (KD) is a model compression technique in which a smaller, less complex model (referred to as the "student") is trained to replicate the behavior of a larger, more complex model (referred to as the "teacher") \cite{hinton2015distilling}. The process involves training the student model to mimic the teacher model's predictions rather than relying on the original training data labels. This approach allows the student model to achieve similar performance levels to the teacher model while significantly reducing computational resources required for inference. 

Traditional model compression techniques, such as pruning and quantization, primarily focus on reducing model size by removing redundant parameters or lowering numerical precision. While effective, these methods may lead to a substantial loss in model accuracy, particularly for complex tasks. In contrast, knowledge distillation leverages the rich information embedded in the teacher model's output to train the student model, providing more nuanced guidance and helping the student model to generalize better. KD also offers significant flexibility in both the size and architecture of the teacher and student models, making it a superior model compression technique. 

The teacher model can be any large, high-accuracy model with no constraints on its architecture, allowing the use of state-of-the-art models. The student model, in contrast, can be smaller and simpler, tailored to the specific constraints of the deployment environment, such as limited memory and computational power. This flexibility enables KD to significantly reduce the student model's size and computational complexity without substantial performance degradation. The architectures of the teacher and student models can differ, with the teacher being a complex model like a state-of-the-art transformer, and the student being a simpler neural network, such as a small transformer with a minimal attention core. This adaptability makes KD highly suitable for various use cases and deployment constraints.

We perform knowledge distillation using a distillation loss to guide the student model to mimic the behaviour of the teacher model, as demonstrated in Equation(\ref{eq:KD_loss}):

\begin{equation} \label{eq:KD_loss}
 L = (\alpha)L_{S} + (1-\alpha)L_{D}
\end{equation}

Here, $\alpha$ is the proportion between student loss (\(L_{S}\)) and distillation loss (\(L_{D}\)). \(L_{S}\) is calculated  from the student predictions matching the raw labels, while \(L_{D}\) measures the match between teacher and student predictions. Various options for distillation loss include KL divergence, MSE, cross-entropy, and cosine similarity \cite{chitty2023survey}. In this study, we use cross-entropy, as it is the most suitable for our model's output.

The final layer of our model is a Convolutional Random Field (CRF) layer, which produces a class prediction as the final output without showing the probability distribution. Therefore, we select categorical cross-entropy (CCE) loss as the distillation loss. Our \(L_{D}\) is calculated using CCE loss between student and teacher class predictions, as shown in Equation(\ref{eq:distillation_loss}):

\begin{equation} \label{eq:distillation_loss}
 L_{D} = -\sum_{i=1}^{N} P_{ti} \times \log(P_{si})
\end{equation}
For \(N\) data points, \(P_{ti}\) is the teacher's categorical prediction and \(P_{si}\) is the student's categorical prediction for the \(i\)\textsuperscript{th} data point.

During training, the student learns from both the hard labels of the training data and the soft targets produced by the teacher model. We use CRF loss as the student loss and CCE loss as the distillation loss, which measures the difference between the student and teacher outputs. The final loss function is a weighted combination of the student and distillation losses, and we experiment with different proportions between these two loss functions. We found that $\alpha$=0.1 provides the highest performance; therefore, the KD results presented in this study all use this $\alpha$ value.

\subsection*{Indoor Localisation Model}
\subsubsection*{Multihead Dual Convolutional Self-Attention (MDCSA), a Transformer-Based Model}

We apply the SoTA Multihead Dual Convolutional Self-Attention (MDCSA) model developed by \cite{jovan2023multimodal} for indoor localisation. The MDCSA model differs from typical transformers by combining convolutional layers with self-attention mechanisms, specifically designed to handle time-series data for indoor localization. This hybrid architecture allows MDCSA to capture both local temporal patterns and long-term dependencies, making it more effective at managing multivariate features and filtering out noise compared to standard transformers.

The MDCSA model includes four main components: Positional Embedding (PE) for transforming RSSI data into spatial and temporal embeddings, Dual Convolutional Self-Attention (DCSA) for combining self-attention with causal convolutions to capture local temporal patterns, Multihead DCSA for learning diverse patterns with varied kernel sizes, and a Conditional Random Field (CRF) layer for ensuring consistent room-level predictions by considering temporal dependencies.

In this study, we aim to develop a TinyML model for indoor localisation by experimenting with two main hyperparameters that determine the model size: hidden size (H) and layer (L). The hidden size refers to the embedding size of all MDCSA components (including PE, and the linear layer inside DCSA components), while the layer represents the number of MDCSA layers. We investigate various MDCSA-based architectures by adjusting these parameters to find an optimal balance between model performance and memory constraints.

\subsubsection*{Mamba-Based Model}
Mamba is a recent class of state space models (SSMs) designed to efficiently and effectively handle sequence modeling by applying a selective mechanism to SSMs \cite{gu2023mamba}. Other architectures like transformers face computational inefficiencies and challenges in modeling long-range dependencies due to their quadratic scaling with sequence length. Mamba addresses these issues by integrating selective state spaces (selective SSM), which combine the strengths of recurrent neural networks (RNNs) and convolutional neural networks (CNNs), to achieve linear scaling in sequence length which is much less than transformer while maintaining high performance across various data modalities, particularly on dense modalities like language \cite{zhang2024mamba} and genomics \cite{guo2024saturn}. Mamba's use of efficient computation strategies, such as kernel fusion and parallel processing, further supports its suitability for real-time processing in resource-constrained environments.Thus, Mamba is suitable to handle the information-dense RSSI data and the lightweight SSM base makes Mamba well-suited for our application in on-device health monitoring systems, where efficient handling of sequential data and computational cost-effectiveness are keys.

\begin{figure}
    \centering
    \includegraphics[width=0.6\linewidth]{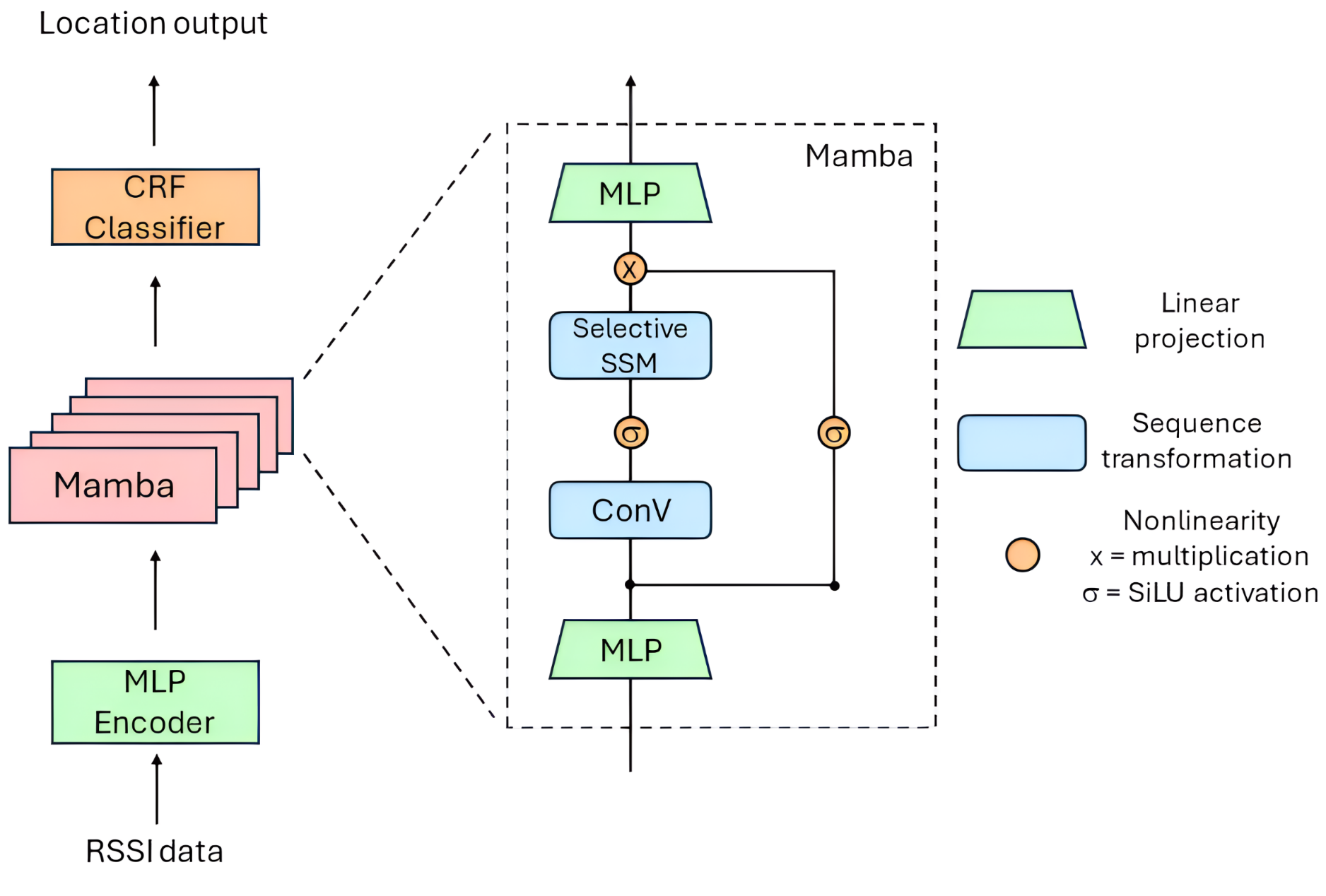}
    \caption{Mamba architecture for RSSI classification}
    \label{fig:mamba_model}
\end{figure}

The key architecture of Mamba consists of three main components: linear projection, sequence transformation, and nonlinearity. Since selective SSMs are standalone sequence transformations, they can be flexibly incorporated into neural networks. Inspired by the Transformer architecture, which typically is an interleaving of linear attention with an MLP (multi-layer perceptron) block, Mamba simplifies this by merging linear projection and sequence transformation into a single Mamba block. These blocks are stacked uniformly, as shown in Figure \ref{fig:mamba_model}. In this study, we use MLPs for the linear projection, convolution and selective SSM for sequence transformation, and SiLU/Swish activation and multiplication for nonlinearity. The number of Mamba layer (blocks) in the stack directly relates to the model size, so we experiment with different numbers of Mamba layers, referring to this parameter as layer (L).

In addition to the basic Mamba block, we incorporate linear embedding and CRF classification layers, as illustrated in Figure \ref{fig:mamba_model}, to enhance the model's performance for localisation tasks. Similar to the MDCSA model, the CRF layer is added at the end of the architecture to perform location classification due to its high capability for sequence labeling tasks. The linear embedding layer maps the raw input data into a hidden latent space with a desired dimension before processing it through the stack of Mamba blocks. This hidden dimension also directly defines the Mamba model size, so we experiment with the appropriate dimension to achieve a practically good performance under the limited constrains. We refer to this parameter as hidden size (H) similar to the MDCSA model.

\subsection*{Experimental Setting}
Our experiments will follow two stages. We will first train an effective indoor localisation model, before modifying it to work within the limited memory constraints available (e.g., via quantization or distillation). 

\subsubsection*{Baseline Models} This experiment involves measuring the performance of different model configurations. We train transformer and Mamba models of various sizes to discover the baseline performances without compression. Each model is trained separately for each dataset and separately for each house of the in-home dataset.

\textbf{MDCSA}: we investigate various MDCSA architectures by adjusting the H and L to observe the model performance and model size. Starting with the original architecture from the referenced paper \cite{jovan2023multimodal}, which uses three layers of kernel sizes 1, 4, and 7 with a hidden size of 256 (referred to as `H256L3'), we then systematically reduce the model size to the smallest possible. We experiment with H = [256, 128, 64, 32, 16, 8, 4, 2] and L = [3, 1]. For clarity, we refer to any model with three layers of kernel sizes 1, 4, and 7 as a `L3 model' and any model with one layer of kernel size 1 as a `L1 model.`

\textbf{Mamba}: We explore different sizes of Mamba by varying the H and the L. We experiment with H = [1, 2, 4, 8, 16, 32, 64, 128], combined with L = [1, 2, 4]. For example, the smallest model consists of a hidden size of 1 and 1 Mamba layer, referred to as `H1L1' following the naming convention. The model performances reach their limit at the hidden size 128 with 4 layers as the performance does not significantly improve much further so we did not extend the model size any larger.

\subsubsection*{Model Compression} This experiment reduces the model size from the baseline model to achieve a suitable model size under the limited memory constraints.

\textbf{Quantization}: We develop TinyML models from the baselines using quantization technique by performing static quantization and dynamic quantization to convert the full models from FP32 precision to int8 precision. We quantize only the linear layers due to the prominence and computational expense of linear layers in Transformer- and Mamba-based models.

\textbf{KD}: For each dataset, we perform KD involving MDCSA and Mamba models, i.e., there are MDCSA teacher and Mamba teacher models for MDCSA student and Mamba student models, for each dataset and each house of the in-home dataset. We identify the most suitable teacher model by selecting the baseline model with the highest performance. Then, we distill the teacher's knowledge to smaller student models derived from the rest of the baseline models.

\subsubsection*{Hybrid Compression Combination} We also investigate the combination of KD and quantization to obtain the benefits of both techniques. KD often leads to improved performance and generalization in the student model and the student can be very tiny as KD is flexible in term of model size. Then quantization can further reduce the size of the tiny student model to significantly decrease memory usage and computational requirements. When combined, these techniques enable the creation of compact, efficient models that maintain high accuracy and performance,  making them well-suited for deployment on resource-constrained devices.

\subsection*{Evaluations}
We evaluate our experimental models using three metrics: macro F1 score (due to the imbalanced nature of the task), accuracy (to assess overall correctness across all predictions), and model size (due to the requirement for the model to fit on the device).

\subsubsection*{Overall F1 Score}
The F1 score measures the model's performance on overall classification, emphasizing the balance between precision and recall. It provides an average class performance across all classes, which is particularly valuable for our in-home dataset where some rooms have less training data and are more challenging to classify, yet they are crucial for accurate localisation (e.g., stairs in a home). Unlike overall accuracy, which may not adequately represent class-specific performance, the F1 score accounts for the importance of these harder-to-classify locations. We compare the F1 score between the baseline and compressed models derived quantization, KD and hybrid. The aim is to achieve an acceptable performance within the device constraints.

\subsubsection*{Accuracy}
Accuracy provides a straightforward measure of how often the system correctly identifies the room, reflecting the overall correctness of predictions. While F1 score offers insights into class-specific performance, accuracy is an intuitive metric that is particularly relevant for real-time applications where consistent correctness is critical. We assess how accuracy changes across baseline and compressed models to evaluate the trade-off between memory constraints and performance. Comparing accuracy and F1 score also highlights the model's ability to balance predictions across frequently and infrequently visited rooms.

\subsubsection*{Model Size}
This metric is essential for developing a TinyML model, as the model needs to fit within the device's limited memory constraints. According to the acceptable memory constraints outlined in Table \ref{tab:TargerMCU_list}, the target sizes are either 64 KB or 32 KB. These constraints provide a good balance between RAM usage and active-mode current, which is important for a long-term indoor monitoring task. Therefore, our goal is to create models that fit within these memory limits. We measure the model size in KB and report the number of parameters. Balancing model size with performance is critical to ensure that the model can be deployed effectively on these resource-constrained devices.

\begin{table}
\footnotesize
    \centering
    \caption{House A: baseline VS compressed models for the best MDCSA and mamba models within the limited constrains. The distillation results came from model that shows the highest baseline validation performance which are H32L4 for mamba (96.59 \% F1 validation, 177 KB, 40248 parameter) and H256L1 for transformer (98.31 \% F1 validation, 10436 KB, 2565148 parameters). The * means model size exceeds memory constraint but the compressed version of the same model is acceptable.}
    \label{tab:mamba_houseA_performance}
    \begin{tabular}{|l|p{0.8cm}|p{.5cm}|p{.5cm}|p{.5cm}|p{.5cm}|p{.5cm}|p{.5cm}|p{.5cm}|p{.5cm}|p{.5cm}|p{.5cm}|p{.5cm}|p{.5cm}|p{.5cm}|p{.5cm}|p{.5cm}|} \hline  
       
        &  number & \multicolumn{3}{c|}{Baseline} & \multicolumn{3}{c|}{Static Quant} & \multicolumn{3}{c|}{Dynamic Quant} & \multicolumn{3}{c|}{Distillation} & \multicolumn{3}{c|}{Distill + Static Quant} \\\hline
 
Model name &  of&F1 & Acc &Size&F1 & Acc &Size & F1 & Acc &Size 
&F1 & Acc &Size&F1 & Acc &Size\\ 

 &   params& (\%) & (\%) &(KB)&(\%)& (\%) &(KB) & (\%) & (\%) &(KB) &(\%)& (\%) &(KB)&(\%)& (\%) &(KB)\\\hline

 Under 64 KB&  & &  && &   && & && &  && & &\\\hline
 
 MDCSA: H16L1&  10588& 84.20*&  97.73
&64*& \textbf{84.36}&  97.57
&44 & 83.15& 97.15
&43
& 82.84*&  97.52
&64*& 82.86& 97.63
&44
\\
 Mamba: H32L1& 10392& 82.71&  \textbf{98.01}
&47& 79.05&  96.59
&26 & 82.18& 97.74
&25
& 77.27&  97.14
&47& 77.32& 97.05
&26
\\\hline
 Under 32 KB &  & &  & & &   & & & & & &  & & & &\\\hline
 MDCSA: H8L1& 2812& 83.33&  97.74
&27& 83.89&  \textbf{97.93}
&26 & \textbf{84.76}& 97.27
&25
& 76.93&  96.04
&27& 76.88& 95.99
&26
\\ Mamba: H8L1&  1432&80.49& 96.44
&12&79.41& 97.27
&12 & 80.46& 97.77
&12
&82.73& 97.76
&12&82.95& 97.78
&12
\\ \hline
        
    \end{tabular}
  
\end{table}

\begin{table*}
\footnotesize
    \centering
    \caption{House B: baseline VS compressed models for the best MDCSA and mamba models within the limited constrains. The distillation results came from model that shows the highest baseline validation performance which are H32L2 for mamba (88.00 \% F1 validation, 92 KB, 20783 parameters) and H256L1 for transformer (94.57  \% F1 validation, 10447 KB, 2567834 parameters). The * means model size exceeds memory constraint but the compressed version of the same model is acceptable.}
    \label{tab:mamba_houseB_performance}
    \begin{tabular}{|l|p{0.8cm}|p{.5cm}|p{.5cm}|p{.5cm}|p{.5cm}|p{.5cm}|p{.5cm}|p{.5cm}|p{.5cm}|p{.5cm}|p{.5cm}|p{.5cm}|p{.5cm}|p{.5cm}|p{.5cm}|p{.5cm}|} \hline  
       
        &  number & \multicolumn{3}{c|}{Baseline} & \multicolumn{3}{c|}{Static Quant} & \multicolumn{3}{c|}{Dynamic Quant} & \multicolumn{3}{c|}{Distillation} & \multicolumn{3}{c|}{Distill + Static Quant} \\\hline
 
Model name &  of&F1 & Acc &Size&F1 & Acc &Size & F1 & Acc &Size 
&F1 & Acc &Size&F1 & Acc &Size\\ 

 &   params& (\%) & (\%) &(KB)&(\%)& (\%) &(KB) & (\%) & (\%) &(KB) &(\%)& (\%) &(KB)&(\%)& (\%) &(KB)\\\hline

  Under 64 KB&  & &  && &   && & && &  && & &\\\hline
   MDCSA: H16L1&  10874& 74.14*&  82.11&65*& 73.84&  \textbf{81.90}&44 & \textbf{74.11}& 81.06
&43
& 74.56*&  82.02&65*& 73.98& 81.68&44
\\Mamba: H16L2& 7263& 72.09&  79.90
&38& 70.27&  78.60
&29 & 70.71& 79.87
&28
& 71.33&  80.19
&38& 71.24& 80.12
&29
\\
\hline
 Under 32 KB&  & &  && &   && & && &  && & &\\\hline
 MDCSA: H8L1&  3018& 67.52&  77.02&28& 67.75&  76.95&26 & 67.11& 76.21
&25
& 69.12&  79.19&28& 69.19& 79.43&26
\\Mamba: H8L1&  1631&72.28& \textbf{80.74}
&12&72.71& 80.71
&13 & 71.73& 79.91
&13
&72.73& 79.94
&12&\textbf{72.79}& 79.96
&13
\\ \hline
        
    \end{tabular}
\end{table*}

\section*{Results and Discussions}

\subsection*{TinyML for In-Home Localisation}
Our investigation into the performance of different SoTA architectures and model compression techniques for in-home localisation under the constraints of TinyML devices revealed several key insights. Overall, the results highlight the robustness of the quantized MDCSA model for in-home localisation under a 64 KB RAM constraint, while a smaller Mamba model is more suitable for environments with even more restrictive memory constraints. Quantization consistently proved to be a valuable technique in reducing model size without sacrificing performance in our task.

The MDCSA model demonstrated the highest performance within the 64 KB RAM limitation. This model achieved a size of 44 KB after quantization with an F1 score ranging from 73.84 to 84.36 across different houses, as presented in Table \ref{tab:mamba_houseA_performance}, \ref{tab:mamba_houseB_performance}, \ref{tab:mamba_houseC_performance} and \ref{tab:mamba_houseD_performance}. The Mamba model, although competitive in terms of both size and performance, was unable to surpass the MDCSA model unless the device constraints were extremely tight, such as less than 32 KB. Under such conditions, the Mamba model outperformed the transformer-based MDCSA model with the performance ranging from F1 score of 72.79 to 83.89 across different houses. These results suggest that the MDCSA model excels in capturing detailed patterns due to its combination of convolution and self-attention through the Multihead DCSA architecture, resulting in an exceptional performance under the acceptable memory constrain. In contrast, the simpler and more compact selective structure state space architecture of the Mamba model is better suited for stricter memory limitations, maintaining reasonable accuracy in highly constrained environments. 

When considering accuracy, we observe that all models achieve higher accuracy than F1 scores across all houses, highlighting the class imbalance in the dataset. This suggests that the models frequently classify majority classes correctly, which boosts the accuracy metric, while they still face challenges in correctly classifying minority classes, as reflected in the lower F1 scores. Such class imbalance is typical for in-home data, as certain areas, like stairs, are less frequently visited in real-world scenarios but are critical for healthcare applications due to their vulnerability.

Additionally, we found that the model with the highest accuracy did not always achieve the highest F1 score. However, models with the highest F1 scores generally achieved accuracy values very close to the highest accuracy model, with differences often less than 1\%. This finding suggests that selecting a model based on its F1 score is more appropriate for this task, given the imbalanced nature of the dataset. Prioritising F1 score ensures better overall performance across all classes, while still maintaining acceptable accuracy.

\begin{table*}
\footnotesize
    \centering
    \caption{House C: baseline VS compressed models for the best MDCSA and mamba models within the limited constrains. The distillation results came from model that shows the highest baseline validation performance which are H64L4 for mamba (88.88 \% F1 validation, 545 KB, 132259 parameters) and H256L1 for transformer (94.79 \% F1 validation, 10445 KB, 2567276 parameters). The * means model size exceed memory constrain but the compressed version of the same model is acceptable.}
    \label{tab:mamba_houseC_performance}
    \begin{tabular}{|l|p{0.8cm}|p{.5cm}|p{.5cm}|p{.5cm}|p{.5cm}|p{.5cm}|p{.5cm}|p{.5cm}|p{.5cm}|p{.5cm}|p{.5cm}|p{.5cm}|p{.5cm}|p{.5cm}|p{.5cm}|p{.5cm}|} \hline  
       
        &  number & \multicolumn{3}{c|}{Baseline} & \multicolumn{3}{c|}{Static Quant} & \multicolumn{3}{c|}{Dynamic Quant} & \multicolumn{3}{c|}{Distillation} & \multicolumn{3}{c|}{Distill + Static Quant} \\\hline
 
Model name &  of&F1 & Acc &Size&F1 & Acc &Size & F1 & Acc &Size 
&F1 & Acc &Size&F1 & Acc &Size\\ 

 &   params& (\%) & (\%) &(KB)&(\%)& (\%) &(KB) & (\%) & (\%) &(KB) &(\%)& (\%) &(KB)&(\%)& (\%) &(KB)\\\hline
 
 Under 64 KB&  & &  && &   && & && &  && & &\\\hline
 MDCSA: H16L1&  10796&83.42*&  94.84
&65*& 83.44&  94.75
&44 & \textbf{83.64}& 94.03
&28
&83.38*&  95.19
&65*& 83.22& \textbf{95.10}
&44
\\ \
 Mamba: H32L1& 10723& 79.12&  93.60
&49& 79.14&  93.62
&26 & 75.46& 93.50
&25
& 77.64&  93.15
&49& 77.75& 93.12
&26
\\\hline

 Under 32 KB&  & &  && &   && & && &  && & &\\\hline
 MDCSA: H8L1&  2956& 72.91&  90.87
&28& 73.20&  91.07
&26 & 73.17& 90.70
&25
& 78.13&  92.45
&28& 78.51& 92.52
&26
\\ Mamba: H8L1&  1571&78.64& 93.37
&12&78.59& 93.35
&13 & 77.06& 93.28
&13
&\textbf{80.60}& 93.44
&12&77.96& \textbf{93.47}
&13
\\ \hline 
        
    \end{tabular}
\end{table*}

\begin{table*}
\footnotesize
    \centering
    \caption{House D: baseline VS compressed models for the best MDCSA and mamba models within the limited constrains. The distillation results came from model that show the highest baseline validation performance which are H64L4 for mamba (89.20 \% F1 validation, 545 KB, 132344 parameters) and H256L1 for transformer (96.62   \% F1 validation, 10445  KB, 2567554 parameters). The * means model size exceeds memory constraint but the compressed version of the same model is acceptable.}
    \label{tab:mamba_houseD_performance}
    \begin{tabular}{|l|p{0.8cm}|p{.5cm}|p{.5cm}|p{.5cm}|p{.5cm}|p{.5cm}|p{.5cm}|p{.5cm}|p{.5cm}|p{.5cm}|p{.5cm}|p{.5cm}|p{.5cm}|p{.5cm}|p{.5cm}|p{.5cm}|} \hline  
       
        &  number & \multicolumn{3}{c|}{Baseline} & \multicolumn{3}{c|}{Static Quant} & \multicolumn{3}{c|}{Dynamic Quant} & \multicolumn{3}{c|}{Distillation} & \multicolumn{3}{c|}{Distill + Static Quant} \\\hline
 
Model name &  of&F1 & Acc &Size&F1 & Acc &Size & F1 & Acc &Size 
&F1 & Acc &Size&F1 & Acc &Size\\ 

 &   params& (\%) & (\%) &(KB)&(\%)& (\%) &(KB) & (\%) & (\%) &(KB) &(\%)& (\%) &(KB)&(\%)& (\%) &(KB)\\\hline

Under 64 KB&  & &  && &   && & && &  && & &\\\hline
 MDCSA: H16L1&  10834& 75.68*&  90.56
&65*& 75.68&  \textbf{90.53}
&44 & \textbf{75.71}& 90.41
&43
& 74.74*&  89.79
&65*& 74.82& 89.84
&44
\\
 Mamba: H32L1& 10776& 71.85&  89.17
&49& 72.33&  89.02
&26 & 69.89& 88.96
&25
& 71.05&  88.53
&49& 70.98& 88.46
&26
\\
\hline
 Under 32 KB&  & &  && &   && & && &  && & &\\\hline
  MDCSA: H8L1&  2986& 63.78&  83.45
&28& 63.55&  83.47
&26 & 64.10& 83.40
&25
& 66.13&  83.42
&28& 66.04& 83.47
&26
\\Mamba: H16L1& 3848& 72.38&  89.12
&21& \textbf{73.17}&  \textbf{89.13}
&16 & 70.03& 89.01
&16
& 72.00&  88.65
&21& 71.96& 88.54
&16
\\ \hline
        
    \end{tabular}
\end{table*}

Selecting the appropriate model and compression technique based on specific memory constraints is crucial for achieving optimal performance in real-world applications. The MDCSA model benefits from its more complex architecture and ability to capture detailed temporal and spatial patterns, which is feasible within a 64 KB constraint. In contrast, the Mamba model, being inherently simpler and more compact, is better suited for environments with stricter memory limitations, where the overhead of the more complex MDCSA model becomes a challenge. 

The quantization process effectively reduced the model size by nearly half while maintaining performance comparable to the baseline. This highlights the efficacy of quantization in deploying models on devices with stringent memory constraints. Conversely, knowledge distillation did not yield significant advantages. Despite expectations that this method would elevate the performance to match that of a larger teacher model, the results showed only marginal improvements or occasional declines in performance, depending on the house. Although the model size is significantly reduced from the teacher model, this technique requires long training times involving training an optimal teacher model and distill the knowledge to a suitable student model, thus it is not an efficient solution in this scenario.

In conclusion, the MDCSA model with a hidden size of 16 and a single layer, along with quantization, is identified as the optimal choice for in-home localisation when the RAM constraint is 64 KB. For devices with a more restrictive 32 KB RAM limitation, a one-layer Mamba model with a hidden size of 8 or 16 is preferable. Model compression does not significantly enhance performance nor considerably reduce the size of the Mamba model, thus it is not considered essential in these scenarios.

\begin{table*}
\footnotesize
    \centering
    \caption{UJIIndoorLoc: baseline VS compressed models for the top performance MDCSA and mamba models within the limited constrains. The distillation results came from model that shows the highest baseline validation performance which are H128L1 for mamba (94.62  \% F1 validation, 783 KB, 194447 parameters) and H256L3 for transformer (97.52   \% F1 validation, 39400 KB, 9802058 parameters).  The * means model size exceed memory constrain but the compressed version of the same model is acceptable.}
    \label{tab:mamba_ujiloc_performance}
    \begin{tabular}{|l|p{0.8cm}|p{.5cm}|p{.5cm}|p{.5cm}|p{.5cm}|p{.5cm}|p{.5cm}|p{.5cm}|p{.5cm}|p{.5cm}|p{.5cm}|p{.5cm}|p{.5cm}|p{.5cm}|p{.5cm}|p{.5cm}|} \hline  
       
        &  number & \multicolumn{3}{c|}{Baseline} & \multicolumn{3}{c|}{Static Quant} & \multicolumn{3}{c|}{Dynamic Quant} & \multicolumn{3}{c|}{Distillation} & \multicolumn{3}{c|}{Distill + Static Quant} \\\hline
 
Model name &  of&F1 & Acc &Size&F1 & Acc &Size & F1 & Acc &Size 
&F1 & Acc &Size&F1 & Acc &Size\\ 

 &   params& (\%) & (\%) &(KB)&(\%)& (\%) &(KB) & (\%) & (\%) &(KB) &(\%)& (\%) &(KB)&(\%)& (\%) &(KB)\\\hline

 Exceed 64 KB & & &  && &   && & && &  && & &\\\hline
 MDCSA: H16L1&  23290& 45.81*&  63.37
&115*& 46.00*
&  19.05
&68* & 45.95*& 63.52
&67*& 48.18*&  65.5
&115*& 48.40*& 18.60
&68*
\\\hline

 Under 64 KB & & &  && &   && & && &  && & &\\\hline
 Mamba: H32L1&  32111&70.02*& 82.97&134*&61.25
& 78.08&58 & 69.8& 77.35&95
&71.51*& 83.65&134*&62.10& 78.49&58
\\
 MDCSA: H8L1&  10978& 19.62&  33.3
&60& 19.63
&  9.42
&45 & 19.40
& 32.97
&44
& 17.49&  30.87
&60& 17.32& 9.78
&45
\\ 
 Mamba: H8L2& 10847& 63.27&  \textbf{79.06} &52& 56.28
&  72.37&40 & 61.45& 72.12&46
& 63.60&  78.49&52& 53.65& 71.72&40
\\
 Mamba: H8L1& 9543& 62.62&  77.57&44& 53.47
&  71.08&31 & 60.5& 70.22&37
& \textbf{64.00}&  78.74&44& 55.19& 72.57&31
\\ \hline
 Under 32 KB & & &  && &   && & && &  && & &\\\hline
 Mamba: H4L1& 6475& \textbf{44.56}&  \textbf{67.21}
&31& 39.88
&  61.96
&27 & 43.76& 61.02
&29
& 41.99&  66.25
&31& 36.14
& 59.77
&27
\\
 Mamba: H2L1& 5020& 9.16&  31.47
&25& 8.90&  29.30
&26& 9.22& 28.89
&25
& 9.79&  31.76
&26
& 9.02& 30.53
&26
\\\hline
        
    \end{tabular}
\end{table*}

\subsection*{TinyML for Large Building Localisation (UJIindoorLOC)}
In the context of UJIindoorLoc, a dataset characterized by a large building and an extensive set of 520 APs, our study shows that the Mamba model performs best within the 64 KB RAM constraint, significantly outperforming other models, which is the opposite to the smaller in-home dataset that the MDCSA outperforms Mamba under this condition.  The most effective Mamba model, utilizing knowledge distillation, achieved a size of 44 KB and an F1 score of 64\% (Table \ref{tab:mamba_ujiloc_performance}). In comparison, the MDCSA model showed notably poorer performance, not only in terms of accuracy but also in terms of size, which exceeded 64 KB even after quantization. This can be attributed to the more challenging dataset, as MDCSA was originally designed for in-home localisation within smaller buildings where the RSSI signal is less complicated due to fewer APs and a smaller building size. 

Given the expansive nature of the data and the large number of APs, models designed for UJIindoorLoc tend to be larger than those optimized for in-home localisation. Consequently, under a stricter 32 KB RAM constraint, these models performed poorly, achieving F1 scores of less than 45 percents. This highlights the challenge of adapting complex models to environments with significant spatial coverage and numerous data points. 

Similar to the house dataset, most models in the UJIIndoorLoc dataset demonstrate higher accuracy than F1 scores, further emphasising the imbalance nature of the task. However, the MDCSA experiences a significant accuracy drop of 20–40\% with static quantization, despite its F1 score remaining relatively stable. This suggests that static quantization disproportionately affects the model's ability to correctly classify majority classes, while its performance on minority classes, which dominate the F1 calculation, remains relatively unaffected. This could be attributed to the sensitivity of transformer-based architectures to precision loss in critical weights or activations that represent majority class distinctions. In contrast, dynamic quantization, which applies quantization during inference rather than pre-quantizing the model weights, shows minimal impact on accuracy. This suggests that dynamic quantization preserves more critical information by adapting the quantization process to runtime requirements.

Unlike the in-home dataset, where quantization proved beneficial, the reduction in model performance is more significant for the UJIindoorLoc dataset because the larger dataset is more complicated and challenging. This suggests that more techniques are required to improve the post-quantization performance, such as fine-tuning the quantize model \cite{hubara2021accurate}, Quantization-Aware Training \cite{jacob2018quantization} or adaptive rounding \cite{nagel2020up}. These techniques are proven effective but require additional time and resources to implement, as they involve retraining or adjusting the model to mitigate the performance loss caused by quantization. While effective, these approaches introduce a trade-off between model complexity and development time, highlighting the need for careful evaluation when targeting highly constrained environments, especially for larger and more complex datasets like UJIindoorLoc.

While quantization did not perform well here, in contrast, knowledge distillation showed a slight improvement in model performance across most model architectures, though with some limitations. The performance enhancements did not match the large teacher models when the model size was significantly decreased. This suggests that while knowledge distillation is effective in transferring knowledge from a larger model to a smaller one, there is a trade-off between model size and performance. The student models were able to capture the general patterns learned by the teacher, but fine-grained details were often lost as the model size decreased. This highlights the need for further refinement in the distillation process, such as incorporating techniques like layer-wise distillation \cite{liang2023less}, to better retain performance in extremely small models. This represents a trade-off, where performance improvements require an extensive time investment in the knowledge distillation process. 

In conclusion, for the UJIindoorLoc dataset, the one-layer Mamba model with a hidden size of 8 and knowledge distillation is the best option under the 64 KB RAM constraint. This model offers a balanced compromise between performance and size, making it suitable for deployment on MCU devices within this constraint. However, developing an efficient TinyML model within the 32 KB RAM constraint for this large dataset remains challenging. The large number of APs demand bigger model sizes to handle high-dimensional inputs, indicating that further model compression or input size reduction may be required to meet such memory constraints without compromising the performance. 

\begin{figure*}
    \centering
    \includegraphics[width=0.7\linewidth]{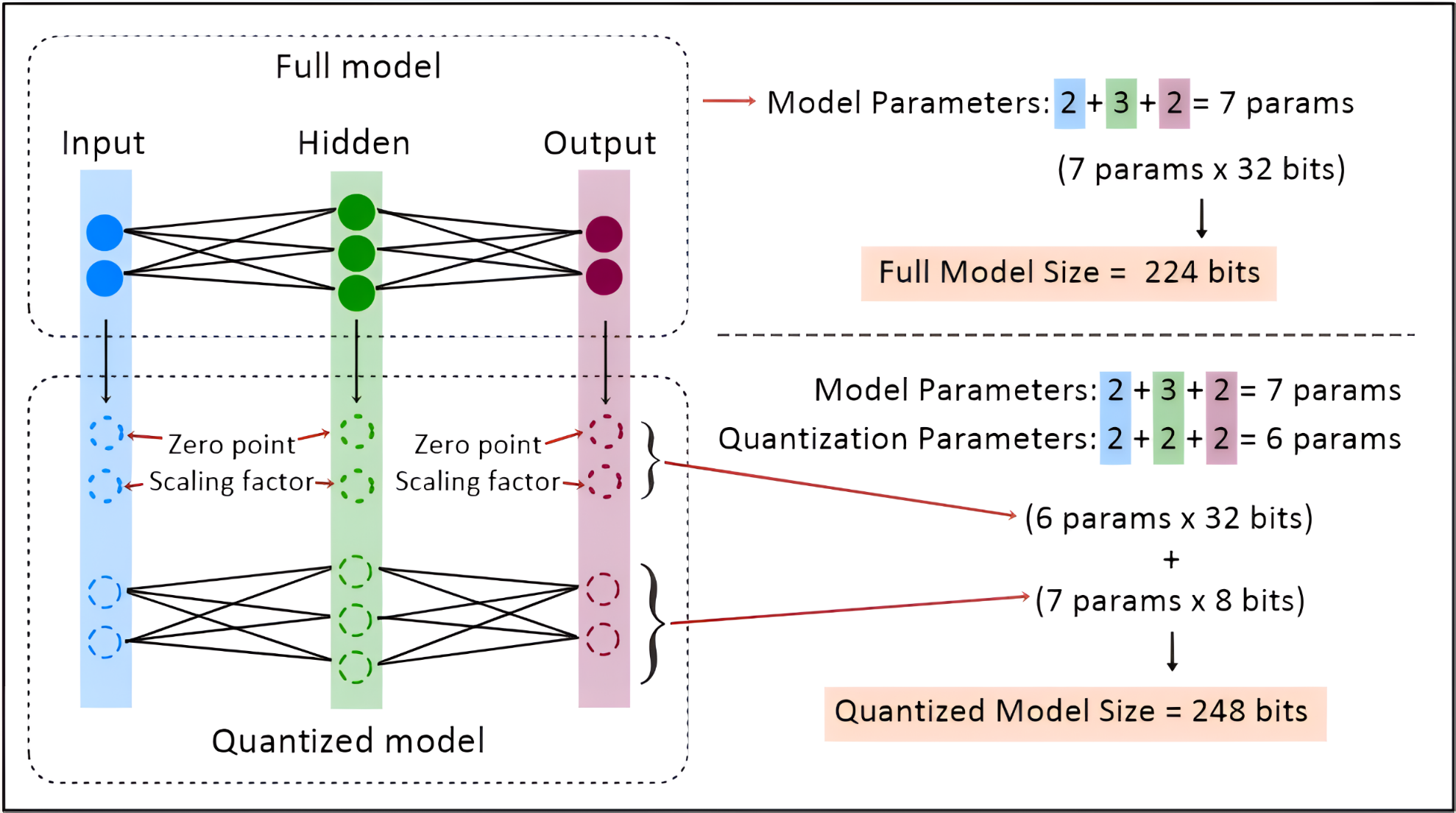}
    \caption{Illustration shows an example case when a quantized model is bigger than the full model due to the overhead from quantization parameters. This usually occurs when quantizing a very small model, indicating the limitation of the quantization method.}
    \label{fig:quant_limit}
\end{figure*}

\subsection*{Limitations of Quantization for TinyML Models}
In this study, we observed the effectiveness and limitations of model quantization in the context of TinyML, specifically focusing on models intended for deployment on devices with highly constrained memory resources for indoor localisation. The findings indicate that quantization, while generally effective in reducing model size for larger models, can be ineffective on a tiny baseline model.

The quantization may not provide size reduction benefits when applied to models that are already very small. For instance, the mamba model of H8L1 in house A has size of 12 KB which remain about the same after quantization. This is because an initial size is close to the minimum required for deployment so there is no significant reduction in size through quantization. In some cases, quantization did not reduce the model size at all or, paradoxically, increased it, as showed in the mamba model of H8L1 in house B and C that the model size increase from 12 KB to 13 KB after quantization. 

This counterintuitive result occurs due to the intrinsic overhead of the quantization process outweigh the benefits of reducing the precision of the model weights. Despite storing the model weights in a compressed int8 format, quantized models still require 32-bit floating point numbers for the scale and zero-point parameters used in dequantization during inference as shown in Figure \ref{fig:quant_limit}. The presence of these 32-bit parameters is necessary to accurately map the lower precision weights back to their original precision range, which is crucial for maintaining the model's performance. However, in very small models, this fixed overhead can represent a significant portion of the total model size, thereby reducing the overall effectiveness of the quantization process.

For TinyML applications where the primary goal is to deploy models on devices with extremely limited memory, these findings underscore the importance of considering the initial model size and the potential overhead introduced by quantization. It highlights that while quantization can be a powerful tool for reducing model size in larger models, its benefits are not universally applicable and may not be suitable for very small models intended for deployment in highly constrained environments.

\subsection*{Limitations of Knowledge distillation for TinyML Models}

KD was explored in this study to reduce model size while maintaining performance, but the results showed only limited improvements. We attribute the suboptimal performance of KD in this study to the simplicity of the distillation method employed. Specifically, our KD approach distilled knowledge only from the final layer of the teacher model, disregarding potentially valuable information from intermediate layers or the relational structure between data samples. The significant size disparity between the teacher model and the student model, limited to 64 KB or less, likely contributed to this limitation, as the distilled knowledge from the teacher's output alone may not have been sufficient to fully train the student model.

More advanced KD methods could address this limitation. Feature-based knowledge distillation \cite{romero2014fitnets}, relation-based knowledge distillation \cite{chen2018darkrank}, and strategies such as auxiliary architectures or adaptive distillation to reduce the performance gap \cite{yang2023categories} are promising approaches. For example, layer-by-layer distillation could enable a more comprehensive transfer of knowledge to the student model, improving its performance.

However, this study identified Mamba-based architectures and quantization as more efficient and computationally less demanding alternatives. Implementing advanced KD methods would require significant computational resources and time, particularly given the computational complexity of layer-wise distillation and the effort involved in training such systems. In contrast, Mamba with quantization offers a practical, efficient solution for TinyML applications.

While advanced KD techniques were not pursued further in this study, their potential for improving performance is acknowledged. Future research will investigate these methods to optimise the knowledge distillation process and enhance the performance of TinyML models.

\subsection*{Complexity and Practical Challenges of Model Deployment}

Deploying machine learning models on MCU-based edge devices presents several complexities and practical challenges that must be carefully addressed. One critical factor is device compatibility, ensuring the model aligns with the hardware limitations of the target MCU, such as memory capacity, processing power, and supported data types (e.g., 8-bit integers or floating-point numbers). Resource constraints are another major challenge, as edge devices typically have limited RAM, Flash, and computational power. 

Another important consideration is the handling of input data. Edge devices may collect data in formats or precision levels that require preprocessing steps to ensure compatibility with the model. For instance, raw RSSI signals may need normalisation or noise reduction before inference. Additionally, the selection of an appropriate deployment framework, such as TensorFlow Lite Micro, ONNX Runtime, or proprietary SDKs, plays a crucial role in translating trained models into formats executable on edge devices. Beyond initial deployment, scalability and maintenance pose challenges, particularly when scaling deployments to multiple devices or updating models in the field to reflect retraining efforts.

To address these challenges, advanced optimisation techniques are crucial for adapting models to resource-constrained environments, as demonstrated in this study, which achieved significant model size reduction while maintaining performance. Profiling and testing are essential to evaluate memory usage, inference time, and power consumption on the target hardware. Developing efficient, lightweight preprocessing pipelines tailored to specific data requirements can further enhance deployment efficiency. Additionally, utilising deployment frameworks like TensorFlow Lite Micro or CMSIS-NN can facilitate seamless integration into edge devices.

While this study primarily focuses on compressing the Mamba and Transformer models for RSSI-based indoor localisation, addressing deployment challenges remains an important next step. Future work will include validating these models on physical edge devices, assessing their performance under real-world operating conditions, and refining deployment strategies to ensure robustness and scalability.

\section*{Conclusion}
In this paper, we presented an in-depth analysis of developing TinyML models for on-device indoor localisation. Our focus was on developing models that fit within the strict memory constraints of low-power MCUs while maintaining high performance and efficiency. With model compression techniques such as quantization and knowledge distillation, we demonstrated that significant reductions in model size are possible without sacrificing, and in some cases improving, localisation performance. Our findings showed that the transformer model, after quantization, provides strong performance under the 64 KB RAM constraint, achieving an optimal balance between model size and indoor localisation performance. Moreover, the Mamba model, designed to be more compact, excelled under even tighter constraints, such as a 32 KB RAM limit, proving its effectiveness in highly resource-limited environments without model compression. The integration of on-device processing further enhances privacy, reduces latency, and lowers operational costs, making it a practical and efficient solution for continuous patient tracking, monitoring, and personalized care. This work provides a framework for converting large SoTA models into tiny models that effectively on highly resource-constrained devices.

\section*{Data Availability}
The In-Home Localisation dataset is available at \byrneurl. \\The UJIIndoorLoc dataset is available at \ujiurl.

\section*{Acknowledgements}
T.S. is supported by the Royal Thai Government scholarship provided by the Ministry of Higher Education, Science, Research and Innovation, Royal Government of Thailand.

\section*{Author contributions statement}
All authors contributed to the study conception and design. T.S. wrote the code, conducted the experiments, analysed the results and prepared all figures. T.S., R.M. and F.J wrote the main manuscript text. R.M and I.C. supervised this project. All authors reviewed the manuscript.

\section*{Additional information}
\textbf{Accession codes}:\myurl\\

\bibliographystyle{ieeetr}
\bibliography{references}

\end{document}